\documentclass[runningheads]{llncs}
\usepackage{graphicx}
\usepackage{booktabs}
\usepackage{url}
\usepackage{tikz}
\usepackage{comment}
\usepackage{amsmath,amssymb}
\usepackage{dsfont}
\usepackage{colortbl}
\usepackage{epsfig}
\usepackage[inline]{enumitem}
\usepackage[skip=0pt]{caption}
\usepackage{soul}
\usepackage[utf8]{inputenc}
\usepackage{listings}
\usepackage[T1]{fontenc}
\usepackage{booktabs}
\usepackage{amsfonts}
\usepackage{bm}
\usepackage{nicefrac}
\usepackage{microtype}
\usepackage{amsmath}
\usepackage{algpseudocode}
\usepackage{algorithm}
\usepackage{numprint}
\usepackage{siunitx}
\usepackage{etoolbox}
\usepackage{cancel}
\newcommand{\ubold}{\fontseries{b}\selectfont}
\robustify\ubold
\usepackage{bbold}
\usepackage{wrapfig}
\usepackage[skip=2pt,font=scriptsize,labelfont=scriptsize]{subcaption}
\usepackage{pifont}
\usepackage{lineno}
\usepackage{array,multirow,graphicx}
\usepackage{booktabs} 
\usepackage{longtable} 
\usepackage[para]{footmisc}
\usepackage{pgfplots}
\usepackage{mathtools}
\pgfplotsset{compat=1.18}
\usepackage[accsupp]{axessibility}  

\usepackage{xcolor}
\usepackage[pagebackref,breaklinks,colorlinks=true,linkcolor=blue,citecolor=blue]{hyperref}

\usepackage{orcidlink}

\begin{document}

\title{SimGraph: A Unified Framework for Scene Graph-Based Image Generation and Editing}

\titlerunning{SimGraph}

\author{Thanh-Nhan Vo\orcidlink{0009-0007-8403-1240}\inst{1,2} 
\and
Trong-Thuan Nguyen\orcidlink{0000-0001-7729-2927}\inst{1,2}$^\spadesuit$ 
\and \\
Tam V. Nguyen\orcidlink{0000-0003-0236-7992}\inst{3} 
\and
Minh-Triet Tran\orcidlink{0000-0003-3046-3041}\inst{1,2}}

\authorrunning{Thanh-Nhan Vo et al.}

\institute{University of Science, VNU-HCM, Vietnam 
\and
Vietnam National University, Ho Chi Minh City, Vietnam
\and
University of Dayton, U.S.A.\\
\email{\{vtnhan,ntthuan\}@selab.hcmus.edu.vn,} \\ 
\email{tmtriet@fit.hcmus.edu.vn}, \email{tamnguyen@udayton.edu}}

\def\thefootnote{$^\spadesuit$}\footnotetext{Corresponding Author}\def\thefootnote{\arabic{footnote}}
\maketitle

\begin{abstract}
Recent advancements in Generative Artificial Intelligence (GenAI) have significantly enhanced the capabilities of both image generation and editing. However, current approaches often treat these tasks separately, leading to inefficiencies and challenges in maintaining spatial consistency and semantic coherence between generated content and edits. Moreover, a major obstacle is the lack of structured control over object relationships and spatial arrangements. Scene graph-based methods, which represent objects and their interrelationships in a structured format, offer a solution by providing greater control over composition and interactions in both image generation and editing. To address this, we introduce SimGraph, a unified framework that integrates scene graph-based image generation and editing, enabling precise control over object interactions, layouts, and spatial coherence. In particular, our framework integrates token-based generation and diffusion-based editing within a single scene graph-driven model, ensuring high-quality and consistent results. Through extensive experiments, we empirically demonstrate that our approach outperforms existing state-of-the-art methods.
\end{abstract}

\section{Introduction}\label{sec:intro}
Recent advances in Generative Artificial Intelligence (GenAI) have revolutionized image creation, demonstrating remarkable capabilities in both generation and editing. In particular, models like Stable Diffusion~\cite{rombach2022high} and DALL·E~\cite{ramesh2021zero} have demonstrated the ability to generate high-quality, diverse images from textual descriptions. By leveraging diffusion-based models and transformers, these models can synthesize realistic images conditioned on natural language prompts, providing users with significant creative freedom. Similarly, image editing has made substantial strides, with models now capable of manipulating images to meet specific criteria, such as altering objects, interactions, backgrounds, or other attributes. Thus, these advances in image generation and editing have broadened the potential applications of GenAI across entertainment, design, and education.

However, despite these impressive developments of GenAI, significant challenges remain in unifying image generation and editing into a single, seamless framework. While text-driven generation models offer flexibility and creativity, they frequently struggle to preserve background details and maintain spatial consistency when edits are introduced, often resulting in artifacts and unrealistic compositions. On the other hand, scene graph-based methods, which utilize structured representations of objects and their relationships, offer enhanced control over spatial layouts and object interactions~\cite{johnson2018image}\cite{yang2022diffusion}. Nevertheless, these methods often require additional fine-tuning or training to handle editing tasks, thereby increasing computational costs and inference time~\cite {li2019pastegan}. Therefore, the current challenge lies in integrating both image generation and editing into a unified framework that is efficient and capable of producing quality, coherent outputs.

In this work, we aim to introduce a unified framework that can seamlessly combine the strengths of both image generation and editing. Existing systems often treat these tasks separately, resulting in inefficiencies, particularly when dealing with complex scene graphs that require precise manipulation. Furthermore, our framework maintains semantic consistency between generated content and modified elements, which remains a persistent issue in current methods~\cite{zhang2024sgedit}\cite{dhamo2020semantic}. By addressing these gaps, our approach aims to provide a more flexible, efficient, and high-quality solution, enabling seamless transitions between generation and editing while ensuring high fidelity and spatial coherence in the final outputs.

Our primary contribution is the introduction of SimGraph, which simultaneously integrates scene graph-based image generation and editing into a single pipeline. Specifically, we propose a unified approach that leverages a single scene graph-driven model to handle both tasks, allowing for fine-grained control over object interactions and spatial relationships. In addition, our framework incorporates token-based generation (Section~\ref{subsec:img_gen}) and diffusion-based editing (Section~\ref{subsec:img_edit}), which are conditioned on scene graph-derived captions and edited prompts to ensure consistency and high-quality results across tasks. Moreover, through extensive experiments, we show that our approach outperforms existing scene graph-based models in image generation~\cite{johnson2018image}\cite{yang2022diffusion}\cite{zhang2024var} and editing tasks~\cite{dhamo2020semantic}\cite{yang2022diffusion}, delivering significant improvements in fidelity, efficiency, and semantic alignment.

The remainder of this paper is organized as follows. Sec.~\ref{sec:related} reviews related work on scene graph-based image generation and editing. In Sec.~\ref{sec:formulation} formulates the problem of integrating image generation and editing from scene graphs. Additionally, Sec.~\ref{sec:approach}, we present the unified framework, detailing the scene graph extraction, token-based image generation, and diffusion-based image editing processes. The experimental setup and results are discussed in Sec.~\ref{sec:experiment}, where we present both quantitative and qualitative evaluations of our approach. Finally, we conclude the paper in Sec.~\ref{sec:conclusion} and discuss potential directions for future research.

\section{Related Work}\label{sec:related}
Scene graphs provide a structured representation of visual scenes by modeling objects as nodes and their interrelationships as edges. In particular, this graph-based approach provides a rich, semantic understanding of the scene, enabling more effective manipulation and control over the visual content~\cite{wang2025indvissgg}\cite{nguyen2025llava}. By explicitly capturing object-level attributes and the spatial and relational context between them, scene graphs have become a powerful tool for a variety of image generation and editing tasks. In addition, scene graphs can be extended to the video domain~\cite{nguyen2024hig}\cite{nguyen2024cyclo}\cite{nguyen2025hyperglm}, enabling dynamic, temporally coherent scene generation and editing. The following subsections investigate the application of scene graphs in both generating realistic images and performing sophisticated image edits.

\subsection{Image Generation from Scene Graphs}\label{rw_ig}
The task of generating images from scene graphs, which represent the relationships between objects in a scene, has attracted considerable attention in recent years due to its potential for controlled, realistic image synthesis. Early works, such as SG2IM~\cite{johnson2018image}, leveraged graph convolutional networks to process scene graphs and generate images by predicting bounding boxes and segmentation masks, which are then refined through cascaded networks. However, previous methods often struggled with the complexities of large graphs and the semantic consistency of object relationships. In response, Herzig et al.~\cite{herzig2020learning} address these limitations by learning canonical graph representations that improve performance with complex scene graphs, providing better robustness to noise and improved generalization across semantically equivalent graphs. In addition, these methods are further complemented by advancements such as PasteGAN~\cite{li2019pastegan}, which introduces a semi-parametric approach for controlling object appearance and interactions via external object crops, ensuring high-quality, semantically consistent results.

Recent progress has also seen the adoption of diffusion models. In particular, SGDiff~\cite{yang2022diffusion} enhances alignment between scene graphs and images by optimizing graph embeddings via masked autoencoding and a contrastive loss. Specifically, SGDiff enables more accurate image generation and manipulation from scene graphs, outperforming earlier methods in image quality and alignment. In addition, SG-Adapter~\cite{shen2024sg} enhances text-to-image generation by incorporating scene graph guidance, which involves multiple objects and relationships. Specifically, SG-Adapter shows the potential to improve alignment between textual descriptions and generated images by refining text embeddings with structured representations.

\subsection{Image Editing from Scene Graphs}\label{rw_ie}
Recent advancements in scene graph-based image generation and manipulation have leveraged the structured, hierarchical nature of scene graphs, which significantly improves image synthesis and editing. SGEdit~\cite{zhang2024sgedit} integrates large language models with generative models, which facilitates precise image modifications at the object level. In particular, SGEdit outperforms existing image editing methods in terms of precision and aesthetics by parsing scene graphs and enabling fine-grained edits through an attention-modulated diffusion model. Similarly, SIMSG~\cite{dhamo2020semantic} focuses on enabling image manipulation by altering scene graph semantics, allowing users to replace objects or alter their relationships. 

In addition, the challenge of complex scene editing is further addressed in SGC-Net~\cite{zhang2022complex}, which improves upon traditional object detection methods by leveraging scene graphs to predict regions of interest and applying a conditional diffusion model to edit them. Moreover, SceneGenie~\cite{farshad2023scenegenie} proposes a scene graph-guided diffusion model to improve text-to-image generation by incorporating geometric information. Furthermore, SGDiff~\cite{yang2022diffusion} directly optimizes the alignment between scene graph embeddings and image features. SGDiff leverages masked autoencoding and contrastive losses to learn representations of scene graphs.

\subsection{Discussion}
\subsubsection{Limitations.}
\begin{figure}[!t]
    \centering
    \includegraphics[width=\linewidth]{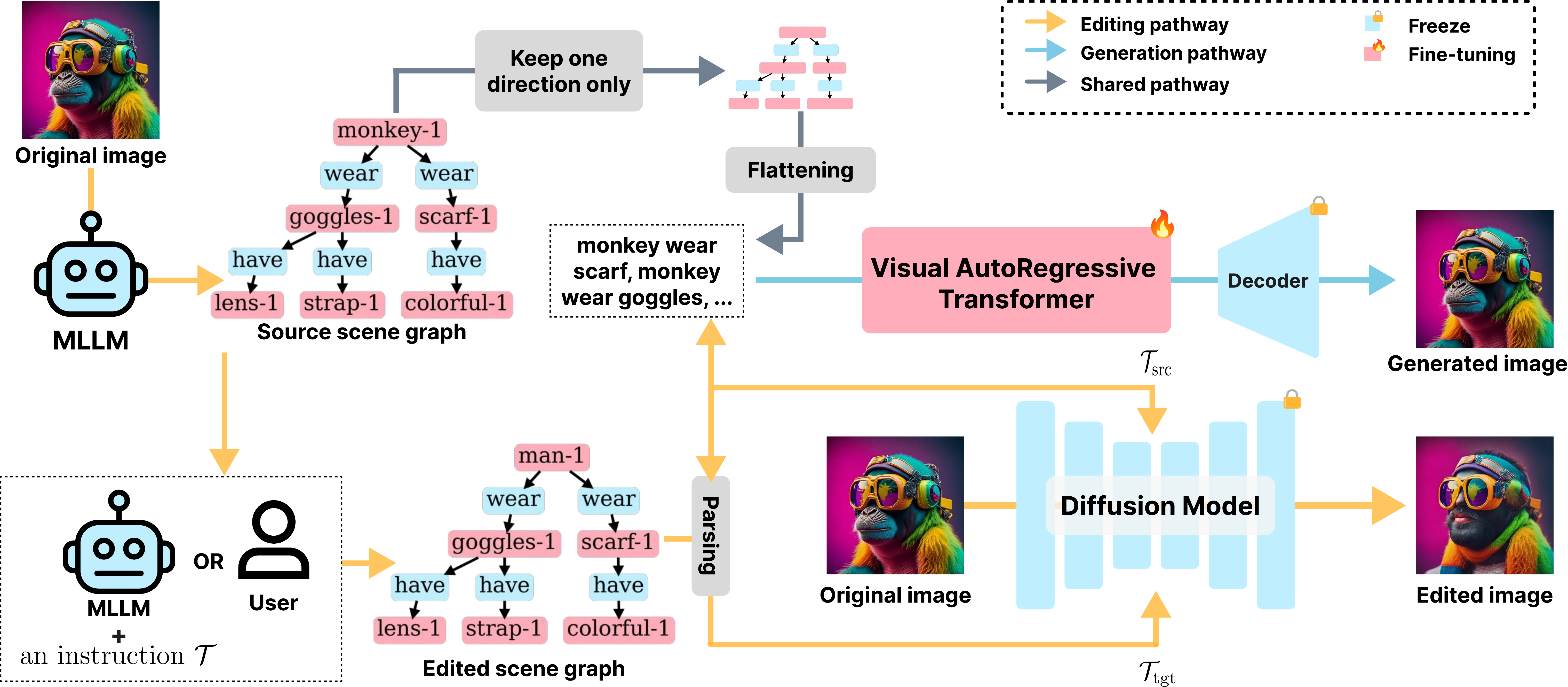}
    \vspace{\baselineskip}
    \caption{Illustration of SimGraph, which shares the same strategy for \textit{scene graph extraction} using MLLM (e.g., Qwen-VL~\cite{bai2023qwen}) (introduced in Sec.~\ref{subsec:sg_extract}). In addition, our framework simultaneously integrates \textit{token-based image generation} (introduced in Sec.~\ref{subsec:img_gen}) and \textit{diffusion model for image editing} (introduced in Sec.~\ref{subsec:img_edit}).}
    \label{fig:our_fw}
\end{figure}

Previous scene graph-based methods for image generation and editing have made significant strides, yet they still face several limitations that our proposed approach aims to address. In image generation, methods like SG2IM~\cite{johnson2018image} and SGDiff~\cite{yang2022diffusion} often struggle to maintain spatial consistency and fine-grained object relationships, especially when handling complex scene graphs. However, these models may generate images with poor alignment between the scene graph's structure and the resulting visual content, leading to artifacts and inaccuracies. 

In image editing, while methods like SGEdit~\cite{zhang2024sgedit} and SIMSG~\cite{dhamo2020semantic} allow for object and relational modifications, they often fail to preserve background details and struggle to achieve seamless transitions between edited and unedited regions. Moreover, these models often require separate pipelines for generation and editing, thereby increasing computational overhead and reducing efficiency. These limitations are noted in Secs.~\ref{rw_ig} and~\ref{rw_ie}, where the challenges of maintaining semantic consistency and improving editing precision are discussed. Thus, our proposed method, as outlined in Secs.~\ref{sec:approach}, addresses these issues by integrating image generation and editing into a unified framework, ensuring semantic coherence.

\subsubsection{Advantages of Our Design.}
The advantages of our proposed framework lie in its unified approach, which leverages a single scene graph-driven model for both generation and editing. By employing token-based generation (Sec.~\ref{subsec:img_gen}) and diffusion-based editing (Sec.~\ref{subsec:img_edit}) in partners, we ensure that both tasks benefit from shared structural control, leading to more consistent and coherent outputs. The integration of joint conditioning on scene graph-derived captions and edited prompts (Alg.~\ref{alg:prompts}) further enhances spatial consistency and enables precise control over object relationships. Additionally, our approach reduces the risk of artifacts and preserves the integrity of background elements during editing, as demonstrated in Alg.~\ref{alg:ledit}. Finally, the ability to process tasks within a unified pipeline reduces computational overhead, providing an efficient solution for real-time applications while maintaining high fidelity and semantic alignment.

\section{Problem Formulation}\label{sec:formulation}

We cast \emph{generation from a scene graph} and \emph{editing guided by scene-graph changes} into a single formulation driven by graph-derived controls. Let $\mathcal{G}$ denote a scene graph (set of relation triplets $t=(s,r,o)$). Specifically, a graph-to-caption transducer $\textsc{S2Cap}(\cdot)$ maps a scene graph to a textual caption $C$ (see Sec.~\ref{subsec:sg_extract}). For editing, given an original graph $\mathcal{G}$ (extracted from an input image $\mathcal{I}$, see Sec.~\ref{subsec:sg_extract}) and an edited graph $\mathcal{G}'$ (obtained from a user instruction), the standard $\textsc{SG2Prompts}(\mathcal{G},\mathcal{G}')$ produces complementary \emph{source} and \emph{target} prompts $(\mathcal{T}_{\text{src}},\mathcal{T}_{\text{tgt}})$ as in Alg.~\ref{alg:prompts}. In particular, we unify both modes (i.e., generation and editing) with a single operator that consumes the caption and (optionally) the prompts and a source image. Formally, we formulate this task as in Eqn.~\eqref{eq:unified_fw}.
\begin{equation}
    \hat{\mathcal{I}}_{\text{final}}
    =
    \mathcal{F}_{\theta}\!\Big(
    \underbrace{\textsc{S2Cap}(\mathcal{G}_{\text{tar}})}_{\text{caption }C\ \text{from Sec.~\ref{subsec:sg_extract}}},\ 
    \underbrace{\mathcal{I}}_{\text{optional source image}},\ 
    \underbrace{\textsc{SG2Prompts}(\mathcal{G},\mathcal{G}')}_{(\mathcal{T}_{\text{src}},\,\mathcal{T}_{\text{tgt}})\ \text{from Alg.~\ref{alg:prompts}}}
    \Big),
    \label{eq:unified_fw}
\end{equation}
where $\theta$ is the model parameter across pathways. In particular, the operator $\mathcal{F}_{\theta}$ selects the appropriate mechanism based on its inputs: (i) \textbf{Generation} (introduced in Sec.~\ref{subsec:img_gen}) when no source image and no prompts are provided $(\mathcal{I}=\varnothing,\ \mathcal{T}_{\text{src}}=\mathcal{T}_{\text{tgt}}=\varnothing)$, reducing to the VAR-based token synthesis conditioned on $C$; (ii) \textbf{Editing} (presented in Sec.~\ref{subsec:img_edit}) when a source image $\mathcal{I}$ and prompts $(\mathcal{T}_{\text{src}},\mathcal{T}_{\text{tgt}})$ are present, invoking the diffusion-based editor under joint conditioning.

\section{Our Proposed Framework}\label{sec:approach}
This section presents SimGraph, a unified framework that simultaneously integrates image generation and editing from a scene graph, as shown in Fig.~\ref{fig:our_fw}.

\subsection{Scene Graph Extraction}\label{subsec:sg_extract}
Given an input image $\mathcal{I}$, we first extract a scene graph using the MLLM (e.g., Qwen-VL-2.5-7B~\cite{bai2023qwen}) that parses visual content and identifies object relations. In particular, each edge is formally represented as a triplet $t_i=(\textit{s}_i,\textit{r}_i,\textit{o}_i)$, where $\textit{r}_i$ is the relation and $\textit{s}_i,\textit{o}_i$ are subject and object, respectively. In addition, triplets are concatenated to form a raw caption $C_{\text{raw}}=t_1,t_2,\dots,t_N$, a textual description.

We then refine $C_{\text{raw}}$ to improve relevance and compactness. First, we prune redundant or bidirectional relations, keeping a single directed instance when both $\textit{s}_i\!\to\!\textit{o}_i$ and $\textit{o}_i\!\to\!\textit{s}_i$ are present. Second, we sort triplets by a salience score $\alpha_i = (w_i^s h_i^s) + (w_i^o h_i^o)$,
where $w_i^s,h_i^s$ and $w_i^o,h_i^o$ are the width and height of subject and object bounding boxes. Specifically, this ordering emphasizes key entities early in the caption and mitigates encoder token bias. As a result, the resulting caption $C$ conditions both the generation and editing pathways.
\begin{algorithm}[t]
\caption{Construct source/target prompts from scene graphs}
\label{alg:prompts}
\begin{algorithmic}[1]
\Require Original scene graph $\mathcal{G}$, edited scene graph $\mathcal{G}'$
\Ensure Source prompt $\mathcal{T}_{\text{src}}$, Target prompt $\mathcal{T}_{\text{tgt}}$
\State $\mathcal{R}(\cdot)$: returns set of relation triplets $(s,r,o)$ from a graph
\State $\textsc{Phrase}(\cdot)$: maps a triplet to a short phrase (e.g., ``dog sitting on bench'')
\Statex
\State $\mathcal{R} \gets \mathcal{R}(\mathcal{G})$; \ $\mathcal{R}' \gets \mathcal{R}(\mathcal{G}')$
\State $\mathcal{R}_{\text{bgd}} \gets \mathcal{R} \cap \mathcal{R}'$ \Comment{stable/background relations}
\State $\mathcal{R}_{\text{new}} \gets \mathcal{R}' \setminus \mathcal{R}$ \Comment{novel/edited relations}
\State $\mathcal{T}_{\mathcal{G}_{\text{bgd}}} \gets \text{concat}\big(\textsc{Phrase}(t)\ \forall\ t \in \mathcal{R}_{\text{bgd}}\big)$
\State $\mathcal{T}_{\mathcal{G}_{\text{new}}} \gets \text{concat}\big(\textsc{Phrase}(t)\ \forall\ t \in \mathcal{R}_{\text{new}}\big)$
\State $\mathcal{T}_{\text{src}} \gets \mathcal{T}_{\mathcal{G}_{\text{bgd}}}$ \Comment{anchors reconstruction of unchanged regions}
\State $\mathcal{T}_{\text{tgt}} \gets \mathcal{T}_{\mathcal{G}_{\text{new}}} \,\Vert\, \mathcal{T}_{\mathcal{G}_{\text{bgd}}}$ \Comment{prioritizes edits while keeping context}
\State \Return $\mathcal{T}_{\text{src}}$, $\mathcal{T}_{\text{tgt}}$
\end{algorithmic}
\end{algorithm}
\subsection{Token-based Image Generation}\label{subsec:img_gen}
For image generation, we synthesize a new image directly from the refined caption $C$ using a Visual AutoRegressive (VAR)~\cite{vo2020visual}. We first obtain a text embedding $\mathbf{e}_t$ from a frozen CLIP encoder~\cite{radford2021learning}. Conditioned on $\mathbf{e}_t$, the VAR predicts a sequence of discrete visual tokens $\mathbf{z}=\{z_1,\dots,z_L\}$, which is formulate as in Eqn.~\eqref{eqn:autoregressive}.
\begin{equation}
    p_{\theta}(\mathbf{z}\mid C)=\prod_{\ell=1}^{L}p_{\theta}\!\left(z_{\ell}\mid z_{<\ell},\mathbf{e}_t\right),
    \label{eqn:autoregressive}
\end{equation}
where $p_{\theta}$ is the conditional autoregressive distribution over tokens parameterized by $\theta$. A frozen VQ-VAE decoder $\psi_v$~\cite{van2017neural} maps the predicted token sequence to pixels, $\hat{I}=\psi_v(\hat{\mathbf{z}})$, ensuring the layout and relations of the image specified by $C$.

\subsection{Diffusion Model for Image Editing}\label{subsec:img_edit}
For editing, we modify a given image $\mathcal{I}$ according to an edited scene graph $\mathcal{G}'$ derived from an instruction $\mathcal{T}$. Similar to image generation (introduced in Sec.~\ref{subsec:img_gen}), we first extract the original scene graph $\mathcal{G}$ as in Sec.~\ref{subsec:sg_extract}, then apply $\mathcal{T}$ to obtain $\mathcal{G}'$. From $\mathcal{G}$ and $\mathcal{G}'$ we construct two complementary prompts (Alg.~\ref{alg:prompts}). 

Let $\mathcal{R}(\cdot)$ return relation triplets and $\textsc{Phrase}(\cdot)$ map triplets to phrases. Specifically, we formally define the stable/background set $\mathcal{R}_{\text{bgd}}=\mathcal{R}(\mathcal{G})\cap\mathcal{R}(\mathcal{G}')$ and novel/edited set $\mathcal{R}_{\text{new}}=\mathcal{R}(\mathcal{G}')\setminus\mathcal{R}(\mathcal{G})$. In addition, the \emph{source} prompt is $\mathcal{T}_{\text{src}}=\text{concat}(\textsc{Phrase}(t):t\in\mathcal{R}_{\text{bgd}})$, specifying content to preserve; the \emph{target} prompt is $\mathcal{T}_{\text{tgt}}=\text{concat}(\textsc{Phrase}(t):t\in\mathcal{R}_{\text{new}})\ \Vert\ \text{concat}(\textsc{Phrase}(t):t\in\mathcal{R}_{\text{bgd}})$, prioritizing edits while retaining essential context. This pairing disentangles preservation from modification, yielding a unified control signal for editing.

We instantiate editing with LEDIT++~\cite{brack2024ledits++} under joint conditioning (Alg.~\ref{alg:ledit}). Given $\mathcal{I}$, we encode to a latent $x_0$ with $\textsc{VAE\_Enc}$, and obtain text conditionings $c_{\text{src}}=\textsc{EncodeTxt}(\mathcal{T}_{\text{src}})$ and $c_{\text{tgt}}=\textsc{EncodeTxt}(\mathcal{T}_{\text{tgt}})$. We first perform latent inversion $\textsc{DDIM\_Invert}(x_0)$ anchored by $\mathcal{T}_{\text{src}}$ to faithfully reconstruct unchanged regions. We then follow a denoising trajectory with joint conditioning and classifier-free guidance (CFG)~\cite{Ho2022ClassifierFreeDG}: at each step, the UNet $f_\theta$ predicts an unconditioned branch and two conditioned branches (source/target), which are blended with weights $w_{\text{src}},w_{\text{tgt}}$ and amplified by guidance scale $s>1$. The source path stabilizes the background structure, while the target path drives the modifications. Finally, $\textsc{VAE\_Dec}$ decodes the refined latent to produce $\mathcal{I}_{\text{edited}}$.

\begin{algorithm}[t]
\caption{Diffusion-based editing with joint source/target conditioning}
\label{alg:ledit}
\begin{algorithmic}[1]
\Require Input image $\mathcal{I}$, source prompt $\mathcal{T}_{\text{src}}$, target prompt $\mathcal{T}_{\text{tgt}}$
\Require Guidance scale $s>1$, weights $w_{\text{src}}, w_{\text{tgt}}\!\in[0,1]$ with $w_{\text{src}}{+}w_{\text{tgt}}{=}1$, diffusion steps $T$
\Ensure Edited image $\mathcal{I}_{\text{edited}}$
\State $\textsc{EncodeTxt}(\cdot)$: text encoder (e.g., CLIP/T5) $\rightarrow$ conditioning embedding
\State $\textnormal{\textsc{VAE\_Enc}}$ / $\textnormal{\textsc{VAE\_Dec}}$: image VAE encoder / decoder
\State $f_\theta(\cdot)$: diffusion UNet predicting noise given latent, timestep, and conditioning
\State $\textnormal{\textsc{DDIM\_Invert}}$, $\textnormal{\textsc{DDIM\_Step}}$: inversion and forward update operators
\Statex
\State $c_{\text{src}} \gets \textsc{EncodeTxt}(\mathcal{T}_{\text{src}})$; \ $c_{\text{tgt}} \gets \textsc{EncodeTxt}(\mathcal{T}_{\text{tgt}})$; \ $c_{\varnothing} \gets \textsc{EncodeTxt}(\text{null})$
\State $x_0 \gets \textnormal{\textsc{VAE\_Enc}}(\mathcal{I})$
\State $x_T \gets \textnormal{\textsc{DDIM\_Invert}}(x_0)$ \Comment{inversion anchored by $\mathcal{T}_{\text{src}}$}
\For{$t = T, T{-}1, \dots, 1$}
    \State $\epsilon_{\varnothing} \gets f_\theta(x_t, t, c_{\varnothing})$
    \State $\epsilon_{\text{src}} \gets f_\theta(x_t, t, c_{\text{src}})$
    \State $\epsilon_{\text{tgt}} \gets f_\theta(x_t, t, c_{\text{tgt}})$
    \State $\epsilon_{\text{text}} \gets w_{\text{src}}\,\epsilon_{\text{src}} + w_{\text{tgt}}\,\epsilon_{\text{tgt}}$
    \State $\epsilon_{\text{pred}} \gets \epsilon_{\varnothing} + s \cdot (\epsilon_{\text{text}} - \epsilon_{\varnothing})$
    \State $x_{t-1} \gets \textnormal{\textsc{DDIM\_Step}}(x_t, \epsilon_{\text{pred}}, t)$
\EndFor
\State $\mathcal{I}_{\text{edited}} \gets \textnormal{\textsc{VAE\_Dec}}(x_0)$
\State \Return $\mathcal{I}_{\text{edited}}$
\end{algorithmic}
\end{algorithm}

\subsection{Loss Function}
Our model with a conditional negative log-likelihood, which is defined in Eqn.~\eqref{eqn:unified_loss}.
\begin{equation}
    \mathcal{L}(\theta) = -\,\mathbb{E}_{(\mathcal{C},\,I^\star)\sim\mathcal{D}}\!\left[\log p_{\theta}\!\big(\,\Phi(I^\star)\ \big|\ \mathcal{C}\,\big)\right],
\label{eqn:unified_loss}
\end{equation}
where $\mathcal{C}$ is the conditioning signal and $\Phi(\cdot)$ the supervision representation. 

For \emph{token-based image generation}, $\mathcal{C}=C$ (the scene-graph caption) and $\Phi=\phi_v$ (the frozen multi-scale VQ-VAE encoder). Therefore, the parameter $p_{\theta}$ is the VAR’s conditional autoregressive distribution over discrete visual tokens. 

For \emph{diffusion-based image editing}, $\mathcal{C}=(\mathcal{T}_{\text{src}},\mathcal{T}_{\text{tgt}})$ (defined in Alg.~\ref{alg:prompts}) and $\Phi$ is the editor’s supervision space. As introduced in Alg.~\ref{alg:prompts}, the parameter $p_{\theta}$ is instantiated by the diffusion model in Alg.~\ref{alg:ledit} and optimized through the standard noise-prediction objective consistent with the conditional likelihood above.

\section{Experiment Results}\label{sec:experiment}
In this section, we evaluate our scene graph–driven framework for image generation and editing. We first summarize implementation details, then report benchmark results using standard metrics. Quantitatively (Tables~\ref{tab:EditVal} and \ref{tab:var-clip-finetune}), our method achieves strong fidelity and competitive accuracy with high efficiency. Additionally, qualitative examples (Figs.~\ref{fig:good}–\ref{fig:failure}) show successful edits and remaining failure modes.

\subsection{Implementation Details}\label{subsec:details}

\noindent\textbf{Model Configuration.} We conduct all experiments on the NVIDIA A100 GPU with 80GB of memory across our proposed framework. Both image generation and image editing share a common step for scene graph extraction, where we utilize Qwen-VL 2.5 (7B)~\cite{qwen2.5-VL} to extract scene graphs from input images. 

For image generation, we employ a pre-trained VAR-CLIP~\cite{zhang2024var}, initialized with a checkpoint based on the VAR-d16 architecture~\cite{zhang2024var}, which integrates CLIP~\cite{radford2021learning} and VQ-VAE~\cite{van2017neural} checkpoints. During fine-tuning, we train with 50 epochs and freeze the weights of the CLIP text encoder and the multi-scale VQ-VAE. Moreover, we use the Adam optimizer with a learning rate of $3 \times 10^{-4} $, $ (\beta_1, \beta_2) = (0.9, 0.95)$, cosine learning rate decay, and gradient clipping at 1.0.

For image editing, we leverage the LEDIT++~\cite{brack2024ledits++}, with Stable Diffusion~\cite{rombach2022high} as the generative backbone to enhance fidelity and editability. In addition, the number of sampling steps is set to 50, and the skip parameter to 25. For editing guidance, we use Qwen-VL to suggest structural modifications. To avoid token truncation in the text encoder, the number of extracted relations is capped at 15. 

\begin{table}[!t]
\centering
\caption{Performance (\%) on EditVal. Accuracy is measured by OwL-ViT~\cite{minderer2022simple}, and fidelity is measured by DINO~\cite{caron2021emerging}. \textit{The best results are highlighted in \textbf{bold}}.}
\vspace{\baselineskip}
\resizebox{.5\textwidth}{!}{%
\begin{tabular}{l|cc}
\toprule
Method & 
Accuracy $\uparrow$ & 
Fidelity $\uparrow$ \\
\midrule
SIMSG~\cite{dhamo2020semantic}  & 0.11 & 0.57 \\
DiffSG~\cite{yang2022diffusion} & 0.01 & 0.13 \\
SimGraph (Ours)                  & \textbf{0.32} & \textbf{0.87} \\
\bottomrule
\end{tabular}%
}
\label{tab:EditVal}
\end{table}

\noindent\textbf{Datasets.} We evaluate our approach using the EditVal~\cite{minderer2022simple} dataset, which is a benchmark specifically designed for the evaluation of image editing methods. Specifically, EditVal comprises images with detailed scene graphs that encode object relationships, enabling a comprehensive analysis of the model's ability to perform edits based on semantic modifications. In addition, the Visual Genome (VG)~\cite{krishna2017visual} and COCO~\cite{lin2014microsoft} subsets of EditVal provide a rich variety of complex scenes and objects, allowing us to test the generalization ability of our framework across different domains. In particular, VG focuses on visual grounding and object relationships, while COCO offers a more diverse set of everyday scenes, ensuring a broad range of editing and generation challenges. By leveraging these datasets, we can assess the performance of our proposed framework in both tasks.

\noindent\textbf{Evaluation Metrics.} We employ standard evaluation metrics for both image generation and image editing tasks. For image generation, we report the FID (Fréchet Inception Distance), which measures the similarity between generated images and real images, with lower values indicating better quality. Additionally, we compute the Inception Score (IS), which evaluates the diversity and quality of generated images, with higher values being desirable. For image editing, we focus on fidelity, using the DINO~\cite{caron2021emerging} metric, which measures how well the edited image preserves the original content while integrating the desired changes. In addition, Accuracy is evaluated using the OwL-ViT~\cite{minderer2022simple} model, which assesses the visual alignment between the original and edited image based on the scene graph.

\subsection{Quantitative Results}\label{subsec:quant}
\begin{table}[!t]
  \centering
  \caption{Performance (\%) on the image generation on Visual Genome and COCO. † COCO results are based on graphs predicted by EGTR~\cite{im2024egtr}; ‡ CLIP scores are computed using graph-derived captions. \textit{The best results are highlighted in \textbf{bold}.}}
  \vspace{\baselineskip}
  \resizebox{.7\textwidth}{!}{%
  \begin{tabular}{l|ccc|cc}
    \toprule
    \multirow{2}{*}{Model} & \multicolumn{3}{c|}{VG} & \multicolumn{2}{c}{COCO} \\
       & FID$\downarrow$ & IS$\uparrow$ & CLIP$\uparrow$ & IS$\uparrow$ & CLIP$\uparrow$ \\ \midrule
    SG2IM~\cite{johnson2018image}                 &  90.5   &  5.5 &  --   &  6.7 &  -- \\
    SGDiff~\cite{yang2022diffusion}                & 26.0    & 16.4 & -- & 17.8 & --\\
    VAR-CLIP~\cite{zhang2024var}   & 56.45 & 16.03 & 18.64 & \textbf{34.54} & \textbf{22.56} \\
    SATURN~\cite{vo2025saturn}       & \textbf{21.62} & \textbf{24.78} & \textbf{21.25}$^{\ddagger}$ & 15.41$^{\dag}$ & 20.98$^{\dag\ddagger}$ \\ \bottomrule
  \end{tabular}%
  }
  \label{tab:var-clip-finetune}
\end{table}

In our experiments, we focus primarily on the image editing process, with image generation results leveraging the pre-existing SATURN model (as detailed in Sec.~\ref{subsec:img_gen}). In particular, the results presented in Table~\ref{tab:EditVal} highlight the strengths of our proposed approach, particularly in terms of fidelity and runtime. Our approach outperforms other scene-graph-based editing methods, such as SGEdit and DiffSG, achieving the highest fidelity score (0.87), which is notably better than SGEdit (0.83) and SIMSG (0.57). This demonstrates our model's ability to preserve fine details while editing, even with modifications to the scene graph. Additionally, our approach offers a significant advantage in runtime, completing edits in 20-30 seconds per image —a marked improvement over SGEdit, which takes 6-10 minutes per image. This efficiency can be attributed to our diffusion-based editing approach (Sec.~\ref{subsec:img_edit}) and the effective use of joint conditioning on source and target prompts (Alg.~\ref{alg:ledit}), enabling accurate image adjustments.

Moreover, the results in Table~\ref{tab:var-clip-finetune} emphasize the robustness of our framework in image generation, where the SATURN model (Sec.~\ref{subsec:img_gen}) contributes to strong FID and IS scores on the VG dataset, outperforming models like SG2IM and SGDiff. In particular, our method achieves the best FID (21.62) and IS (24.78) scores, demonstrating a high-quality generation process prior to the image editing step. Notably, the CLIP score for the COCO dataset (20.98) is also competitive, indicating that our approach generates images that align well with semantic textual representations. Therefore, these results highlight the synergy between the generation and editing phases, with scene graphs serving as a unified control interface for both tasks. Overall, our approach demonstrates superior fidelity, efficiency, and performance compared to existing scene-graph-based methods.

\subsection{Qualitative Results}\label{subsec:quanl}
\begin{figure}[!t]
    \centering
    \includegraphics[width=\linewidth]{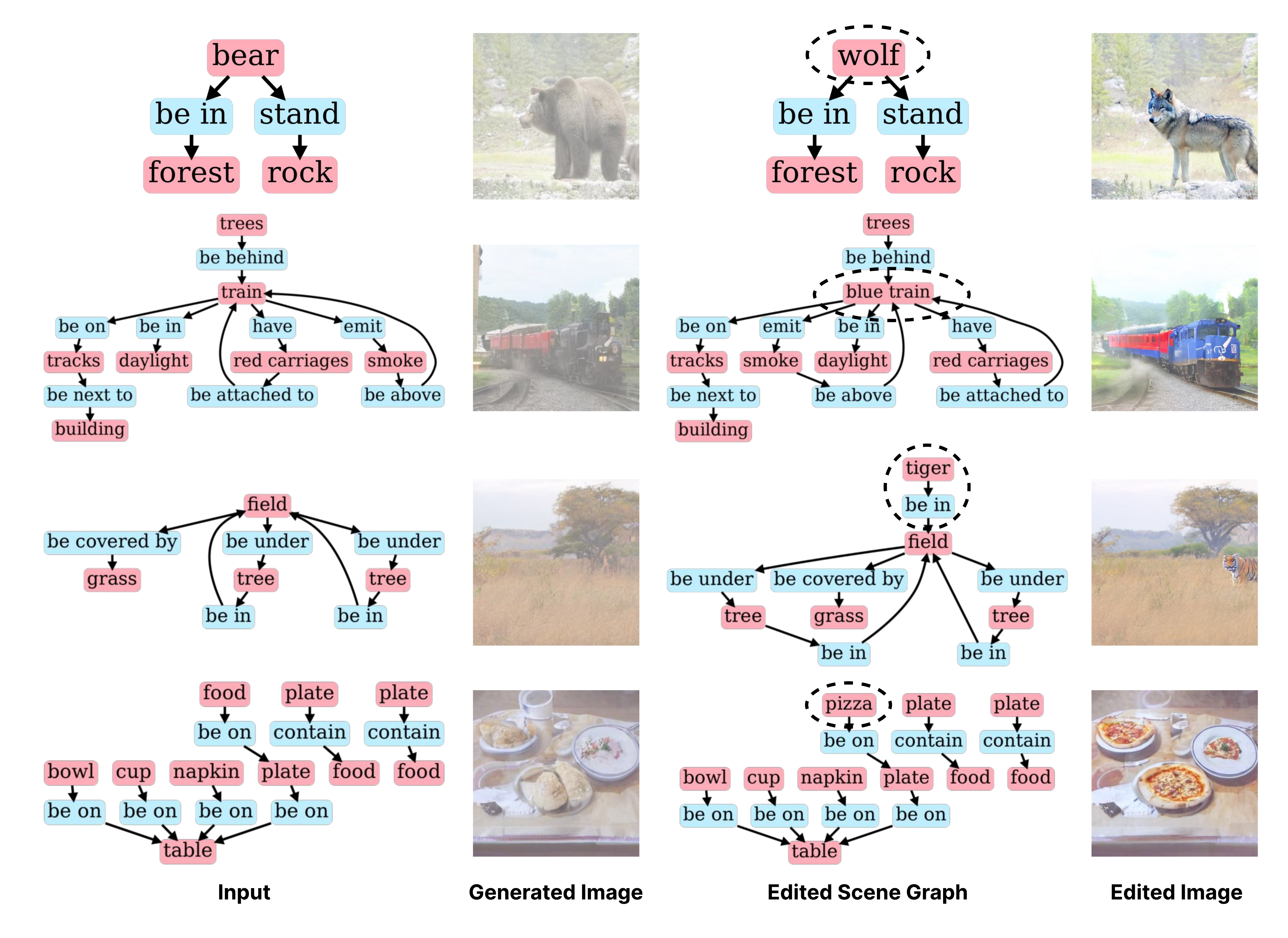}
    \vspace{\baselineskip}
    \caption{Illustration of image generation and editing from scene graphs using our framework. The left side shows the input image with its corresponding scene graph. The generated image on the right demonstrates the model's ability to faithfully recreate the scene from the extracted scene graph. The edited scene graph highlights the modifications made, such as replacing the ``bear'' with a ``wolf'' in the forest and adding a ``tiger''. \textit{Best viewed in color and zoomed in.}}
    \label{fig:good}
\end{figure}

Fig.~\ref{fig:good} demonstrates the effectiveness of our proposed framework (as outlined in Sec.~\ref{sec:approach}) in both image generation and editing from scene graphs. In the generated images, the model successfully captures the spatial and relational semantics encoded in the scene graph, such as ``bear be in forest'' and ``trees be behind train'', producing images that align with these relationships (see Sec.~\ref{subsec:img_gen} for generation details). In the edited images, our framework enables precise modifications, as evidenced by the replacement of the``bear'' with the ``wolf'' in the forest scene and the addition of the ``tiger'' in the field. In particular, these modifications are driven by the edited scene graph, which illustrates how the relationships between objects are altered (refer to Alg.~\ref{alg:prompts} for the construction of source and target prompts). In addition, this directly influences the generated visual content, showcasing the flexibility and control of our approach. Moreover, the ability to edit specific elements, such as changing animals or backgrounds, highlights the power of scene graph-driven image manipulation, ensuring that the image maintains visual coherence while allowing meaningful edits (defined in Sec.~\ref {subsec:img_edit}).

\begin{figure}[!t]
    \centering
    \includegraphics[width=\linewidth]{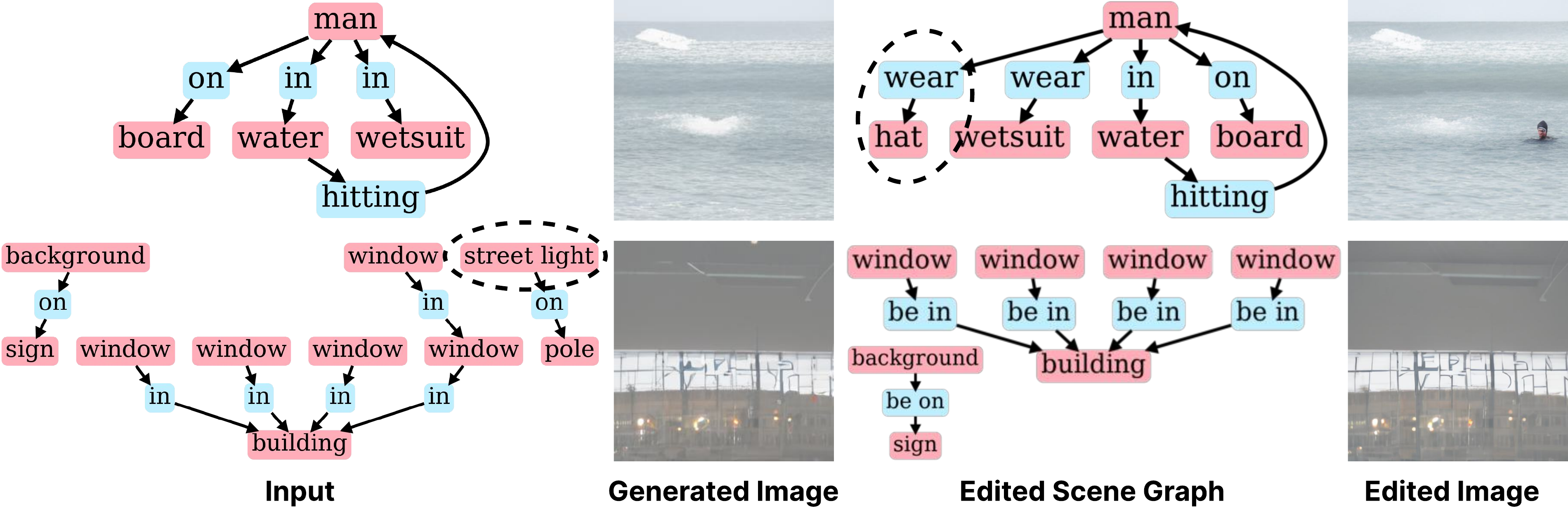}
    \vspace{\baselineskip}
    \caption{Illustration of failure cases. \textit{Best viewed in color and zoomed in.}}
    \label{fig:failure}
\end{figure}

The failure cases shown in Fig.~\ref{fig:failure} highlight challenges in the proposed framework's ability to accurately generate and edit images when complex scene graph modifications are involved. In the image generation phase, while the model produces an image based on the initial scene graph (e.g., ``man on board in water''), the edited scene graph (e.g., changing ``man in wetsuit'' to ``man in hat'') leads to an edited image that fails to reflect the desired change accurately. The model struggles with manipulating objects in the scene, particularly when multiple elements are involved (e.g., changing the man’s outfit while keeping other objects, such as the board and water, intact). Additionally, during background editing, the framework fails to effectively adapt to the new relationships among windows, signs, and the building, leading to a mismatch between the intended scene graph modifications and the final image output. This reveals the model's limitations in handling certain complex relationships and highlights the need to improve preservation of spatial consistency and the accurate application of modifications to foreground and background objects during the editing process.

\section{Conclusion}\label{sec:conclusion}
In this paper, we have presented SimGraph, a unified framework for image generation and editing based on scene graphs. By integrating scene graph-based token generation and diffusion-based editing, we have developed a model that ensures consistency and high-quality results across both tasks. Our approach utilizes a single scene graph-driven model for both image generation and editing, enabling fine-grained control over object interactions and spatial arrangements. Through extensive experiments, we demonstrated that our proposed approach outperforms existing scene graph-based models in fidelity and efficiency.

While our proposed framework offers a promising solution for integrating image generation and editing, there remain avenues for improvement and future exploration. One area for future work is enhancing the robustness of our proposed framework to handle more complex, dynamic scenes, such as those with multiple objects undergoing simultaneous edits. Furthermore, additional research could investigate the incorporation of multimodal inputs, such as the simultaneous combination of textual and visual cues, to facilitate even more intuitive and flexible user interactions. Moreover, it reduces computational overhead while maintaining high output quality, particularly for real-time applications. 

\vspace{5mm}
\textbf{Acknowledgment}. This research is funded by Vietnam National University HoChiMinh City (VNU-HCM) under a project within the framework of the Program titled ``Strengthening the capacity for education and basic scientific research integrated with strategic technologies at VNU-HCM, aiming to achieve advanced standards comparable to regional and global levels during the 2025-2030 period, with a vision toward 2045''

\newpage
\bibliographystyle{splncs04}
\bibliography{main}

@article{zhang2024var,
  title={Var-clip: Text-to-image generator with visual auto-regressive modeling},
  author={Zhang, Qian and Dai, Xiangzi and Yang, Ninghua and An, Xiang and Feng, Ziyong and Ren, Xingyu},
  journal={arXiv preprint arXiv:2408.01181},
  year={2024}
}

@inproceedings{im2024egtr,
  title={Egtr: Extracting graph from transformer for scene graph generation},
  author={Im, Jinbae and Nam, JeongYeon and Park, Nokyung and Lee, Hyungmin and Park, Seunghyun},
  booktitle={Proceedings of the IEEE/CVF Conference on Computer Vision and Pattern Recognition},
  pages={24229--24238},
  year={2024}
}

@article{krishna2017visual,
  title={Visual genome: Connecting language and vision using crowdsourced dense image annotations},
  author={Krishna, Ranjay and Zhu, Yuke and Groth, Oliver and Johnson, Justin and Hata, Kenji and Kravitz, Joshua and Chen, Stephanie and Kalantidis, Yannis and Li, Li-Jia and Shamma, David A and others},
  journal={International journal of computer vision},
  volume={123},
  pages={32--73},
  year={2017},
  publisher={Springer}
}

@inproceedings{rombach2022high,
  title={High-resolution image synthesis with latent diffusion models},
  author={Rombach, Robin and Blattmann, Andreas and Lorenz, Dominik and Esser, Patrick and Ommer, Bj{\"o}rn},
  booktitle={Proceedings of the IEEE/CVF conference on computer vision and pattern recognition},
  pages={10684--10695},
  year={2022}
}

@inproceedings{brack2024ledits++,
  title={Ledits++: Limitless image editing using text-to-image models},
  author={Brack, Manuel and Friedrich, Felix and Kornmeier, Katharia and Tsaban, Linoy and Schramowski, Patrick and Kersting, Kristian and Passos, Apolin{\'a}rio},
  booktitle={Proceedings of the IEEE/CVF conference on computer vision and pattern recognition},
  pages={8861--8870},
  year={2024}
}

@inproceedings{johnson2018image,
  title={Image generation from scene graphs},
  author={Johnson, Justin and Gupta, Agrim and Fei-Fei, Li},
  booktitle={Proceedings of the IEEE conference on computer vision and pattern recognition},
  pages={1219--1228},
  year={2018}
}

@article{zhang2024sgedit,
  title={Sgedit: Bridging llm with text2image generative model for scene graph-based image editing},
  author={Zhang, Zhiyuan and Chen, DongDong and Liao, Jing},
  journal={arXiv preprint arXiv:2410.11815},
  year={2024}
}

@article{shen2024sg,
  title={Sg-adapter: Enhancing text-to-image generation with scene graph guidance},
  author={Shen, Guibao and Wang, Luozhou and Lin, Jiantao and Ge, Wenhang and Zhang, Chaozhe and Tao, Xin and Zhang, Yuan and Wan, Pengfei and Wang, Zhongyuan and Chen, Guangyong and others},
  journal={arXiv preprint arXiv:2405.15321},
  year={2024}
}

@article{Ho2022ClassifierFreeDG,
  title={Classifier-Free Diffusion Guidance},
  author={Jonathan Ho},
  journal={ArXiv},
  year={2022},
  volume={abs/2207.12598},
  url={https://api.semanticscholar.org/CorpusID:249145348}
}

@misc{qwen2.5-VL,
    title = {Qwen2.5-VL},
    url = {https://qwenlm.github.io/blog/qwen2.5-vl/},
    author = {Qwen Team},
    month = {January},
    year = {2025}
}

@inproceedings{vo2025saturn,
  title={SATURN: Autoregressive Image Generation Guided by Scene Graphs},
  author={Vo, Thanh-Nhan and Nguyen, Trong-Thuan and Nguyen, Tam V and Tran, Minh-Triet},
  booktitle={2025 International Conference on Multimedia Analysis and Pattern Recognition (MAPR)},
  pages={1--6},
  year={2025},
  organization={IEEE}
}

@inproceedings{dhamo2020semantic,
  title={Semantic image manipulation using scene graphs},
  author={Dhamo, Helisa and Farshad, Azade and Laina, Iro and Navab, Nassir and Hager, Gregory D and Tombari, Federico and Rupprecht, Christian},
  booktitle={Proceedings of the IEEE/CVF conference on computer vision and pattern recognition},
  pages={5213--5222},
  year={2020}
}

@inproceedings{minderer2022simple,
  title={Simple open-vocabulary object detection},
  author={Minderer, Matthias and Gritsenko, Alexey and Stone, Austin and Neumann, Maxim and Weissenborn, Dirk and Dosovitskiy, Alexey and Mahendran, Aravindh and Arnab, Anurag and Dehghani, Mostafa and Shen, Zhuoran and others},
  booktitle={European conference on computer vision},
  pages={728--755},
  year={2022},
  organization={Springer}
}

@inproceedings{caron2021emerging,
  title={Emerging properties in self-supervised vision transformers},
  author={Caron, Mathilde and Touvron, Hugo and Misra, Ishan and J{\'e}gou, Herv{\'e} and Mairal, Julien and Bojanowski, Piotr and Joulin, Armand},
  booktitle={Proceedings of the IEEE/CVF international conference on computer vision},
  pages={9650--9660},
  year={2021}
}

@inproceedings{lin2014microsoft,
  title={Microsoft coco: Common objects in context},
  author={Lin, Tsung-Yi and Maire, Michael and Belongie, Serge and Hays, James and Perona, Pietro and Ramanan, Deva and Doll{\'a}r, Piotr and Zitnick, C Lawrence},
  booktitle={European conference on computer vision},
  pages={740--755},
  year={2014},
  organization={Springer}
}

@article{bai2023qwen,
  title={Qwen technical report},
  author={Bai, Jinze and Bai, Shuai and Chu, Yunfei and Cui, Zeyu and Dang, Kai and Deng, Xiaodong and Fan, Yang and Ge, Wenbin and Han, Yu and Huang, Fei and others},
  journal={arXiv preprint arXiv:2309.16609},
  year={2023}
}

@article{van2017neural,
  title={Neural discrete representation learning},
  author={Van Den Oord, Aaron and Vinyals, Oriol and others},
  journal={Advances in neural information processing systems},
  volume={30},
  year={2017}
}

@inproceedings{radford2021learning,
  title={Learning transferable visual models from natural language supervision},
  author={Radford, Alec and Kim, Jong Wook and Hallacy, Chris and Ramesh, Aditya and Goh, Gabriel and Agarwal, Sandhini and Sastry, Girish and Askell, Amanda and Mishkin, Pamela and Clark, Jack and others},
  booktitle={International conference on machine learning},
  pages={8748--8763},
  year={2021},
  organization={PmLR}
}

@inproceedings{vo2020visual,
  title={Visual-relation conscious image generation from structured-text},
  author={Vo, Duc Minh and Sugimoto, Akihiro},
  booktitle={European conference on computer vision},
  pages={290--306},
  year={2020},
  organization={Springer}
}

@article{yang2022diffusion,
  title={Diffusion-based scene graph to image generation with masked contrastive pre-training},
  author={Yang, Ling and Huang, Zhilin and Song, Yang and Hong, Shenda and Li, Guohao and Zhang, Wentao and Cui, Bin and Ghanem, Bernard and Yang, Ming-Hsuan},
  journal={arXiv preprint arXiv:2211.11138},
  year={2022}
}

@inproceedings{nguyen2024hig,
  title={Hig: Hierarchical interlacement graph approach to scene graph generation in video understanding},
  author={Nguyen, Trong-Thuan and Nguyen, Pha and Luu, Khoa},
  booktitle={Proceedings of the IEEE/CVF Conference on Computer Vision and Pattern Recognition},
  pages={18384--18394},
  year={2024}
}

@article{nguyen2024cyclo,
  title={CYCLO: Cyclic graph transformer approach to multi-object relationship modeling in aerial videos},
  author={Nguyen, Trong-Thuan and Nguyen, Pha and Li, Xin and Cothren, Jackson and Yilmaz, Alper and Luu, Khoa},
  journal={Advances in Neural Information Processing Systems},
  volume={37},
  pages={90355--90383},
  year={2024}
}

@inproceedings{nguyen2025hyperglm,
  title={Hyperglm: Hypergraph for video scene graph generation and anticipation},
  author={Nguyen, Trong-Thuan and Nguyen, Pha and Cothren, Jackson and Yilmaz, Alper and Luu, Khoa},
  booktitle={Proceedings of the Computer Vision and Pattern Recognition Conference},
  pages={29150--29160},
  year={2025}
}

@inproceedings{ramesh2021zero,
  title={Zero-shot text-to-image generation},
  author={Ramesh, Aditya and Pavlov, Mikhail and Goh, Gabriel and Gray, Scott and Voss, Chelsea and Radford, Alec and Chen, Mark and Sutskever, Ilya},
  booktitle={International conference on machine learning},
  pages={8821--8831},
  year={2021},
  organization={Pmlr}
}

@inproceedings{nguyen2025llava,
  title={LLaVA-SNIPPER: Scene-Graph-based Inference with Multimodal LLMs for Explainable Out-of-Context Misinformation},
  author={Nguyen, Trong-Thuan and Tran, Minh-Triet},
  booktitle={Proceedings of the 2nd Workshop on Security-Centric Strategies for Combating Information Disorder},
  pages={1--10},
  year={2025}
}

@article{wang2025indvissgg,
  title={IndVisSGG: VLM-based scene graph generation for industrial spatial intelligence},
  author={Wang, Zuoxu and Yan, Zhijie and Li, Shufei and Liu, Jihong},
  journal={Advanced Engineering Informatics},
  volume={65},
  pages={103107},
  year={2025},
  publisher={Elsevier}
}

@inproceedings{herzig2020learning,
  title={Learning canonical representations for scene graph to image generation},
  author={Herzig, Roei and Bar, Amir and Xu, Huijuan and Chechik, Gal and Darrell, Trevor and Globerson, Amir},
  booktitle={European Conference on Computer Vision},
  pages={210--227},
  year={2020},
  organization={Springer}
}

@article{li2019pastegan,
  title={Pastegan: A semi-parametric method to generate image from scene graph},
  author={Li, Yikang and Ma, Tao and Bai, Yeqi and Duan, Nan and Wei, Sining and Wang, Xiaogang},
  journal={Advances in Neural Information Processing Systems},
  volume={32},
  year={2019}
}

@article{zhang2022complex,
  title={Complex scene image editing by scene graph comprehension},
  author={Zhang, Zhongping and He, Huiwen and Plummer, Bryan A and Liao, Zhenyu and Wang, Huayan},
  journal={arXiv preprint arXiv:2203.12849},
  year={2022}
}

@inproceedings{farshad2023scenegenie,
  title={Scenegenie: Scene graph guided diffusion models for image synthesis},
  author={Farshad, Azade and Yeganeh, Yousef and Chi, Yu and Shen, Chengzhi and Ommer, B{\"o}jrn and Navab, Nassir},
  booktitle={Proceedings of the IEEE/CVF International Conference on Computer Vision},
  pages={88--98},
  year={2023}
}

\end{document}